\documentclass[10pt,twocolumn,letterpaper]{article}

\usepackage{iccv}
\usepackage{times}
\usepackage{epsfig}
\usepackage{graphicx}
\usepackage{amsmath}
\usepackage{amssymb}

\usepackage{enumitem}
\usepackage{booktabs}

\usepackage[pagebackref=true,breaklinks=true,letterpaper=true,colorlinks,bookmarks=false]{hyperref}

\iccvfinalcopy 

\def\iccvPaperID{3624} 

\ificcvfinal\fi

\begin{document}

\title{Identity Encoder for Personalized Diffusion}

\author{
Yu-Chuan Su \and
Kelvin C.K. Chan \and
Yandong Li \and
Yang Zhao \and
Han Zhang \and
Boqing Gong \and
Huisheng Wang \and
Xuhui Jia}

\maketitle
\ificcvfinal\thispagestyle{empty}\fi

\begin{abstract}
Many applications can benefit from personalized image generation models, including image enhancement, video conferences, just to name a few. Existing works achieved personalization by fine-tuning one model for each person. While being successful, this approach incurs additional computation and storage overhead for each new identity. Furthermore, it usually expects tens or hundreds of examples per identity to achieve the best performance. To overcome these challenges, we propose an encoder-based approach for personalization. We learn an identity encoder which can extract an identity representation from a set of reference images of a subject, together with a diffusion generator that can generate new images of the subject conditioned on the identity representation. Once being trained, the model can be used to generate images of arbitrary identities given a few examples even if the model hasn't been trained on the identity. Our approach greatly reduces the overhead for personalized image generation and is more applicable in many potential applications. Empirical results show that our approach consistently outperforms existing fine-tuning based approach in both image generation and reconstruction, and the outputs is preferred by users more than $95\%$ of the time compared with the best performing baseline.
\end{abstract}

\section{Introduction}
\label{sec:intro}
\begin{figure}[t]
    \centering
    \includegraphics[width=\linewidth]{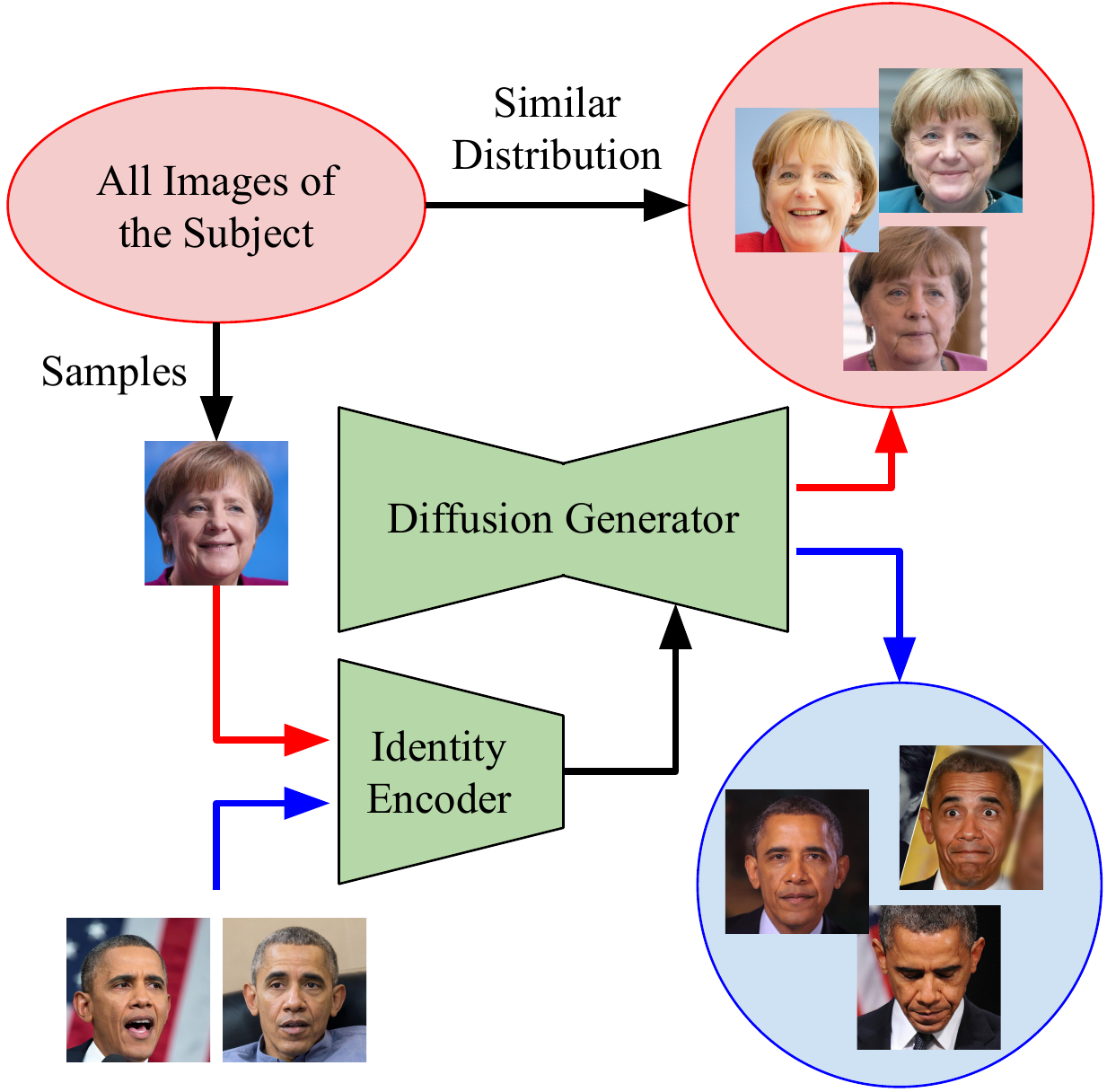}
    \caption{
        Given reference images of a subject, our goal is to learn an identity encoder and a diffusion generator such that the identity encoder can extract an identity representation from the references, and the diffusion generator can generate diverse images of the subject conditioning on the representation. The same model can be used to generate images of different subjects given different references, and the identity encoder can extracts a high-quality representation from just a few or even a single image.
    }
    \vspace{-9pt}
    \label{fig:concept}
\end{figure}

Recent years have seen a tremendous success in the task of face generation. The emerge of VAEs~\cite{van2017neural,esser2021taming}, GANs~\cite{brock2018large,karras2018progressive,karras2019style,karras2020analyzing,karras2021alias}, and diffusion models~\cite{ho2020denoising,rombach2022high} has enabled unlimited possibility in synthesizing faces with various content given different forms of conditions. 
However, models trained on generic facial images are often incapable of generating images of a specific identity, which is a highly desirable feature in practical uses. In order to achieve this, the model must understand the appearance of the desired identity, which are likely not included in the training set. Therefore, a prevalent approach~\cite{nitzan2022mystyle,gal2022image,ruiz2023dreambooth,mokady2022null} is to fine-tune a pre-trained synthesis model using multiple images capturing the corresponding face. The fine-tuned models have been shown capable of synthesizing \textit{personalized} images with diverse content.

Despite the simplicity and remarkable performance, the applicability of the aforementioned fine-tuning paradigm is inevitably limited by the need of lengthy optimization process. Specifically, these methods require model fine-tuning for each identity, which takes minutes to hours to complete. As a result, they are infeasible in cases that require fast response, such as interactive editing. More importantly, they often require up to hundreds of images for fine-tuning due to the identity-agnostic nature of the pre-trained models. Furthermore, the storage cost increases with the number of identities, essentially prohibiting the methods to generalize to arbitrary identities. Therefore, a framework that does not require identity-specific fine-tuning is necessary. 

We take the first step to explore the possibility of developing a feed-forward framework for \textit{personalized face generation}. Contrary to existing methods that obtain an identity-specific embedding through prompt and model tuning, we employ a learnable encoder to generate identity embeddings from input images provided by the user. See Fig.~\ref{fig:concept}. Our method, termed identity encoder, obtains an embedding without any optimization, achieving up to 29 times speedup over existing methods.

While our method is conceptually simple, the acquisition of an effective encoder remains formidable. Specifically, directly training the encoder and diffusion model jointly leads to degenerated outputs, resulting in images with no variations. In this work, we explore the keys in training to achieve balance between \textit{identity preservation} and \textit{output diversity}. 
First, during training, we impose an \textit{identity-preservation} constraint on the learned embeddings, in which the convex hull formed by the embeddings from the same person must represent the same identity. This is essential in forcing the model to learn an identity-specific embedding that preserve variations among outputs.
Second, to further improve identity preservation, we introduce a \textit{soft-nearest neighbor identity loss}, which localizes the embeddings obtained from the same identity. 
Third, to alleviate the lack of identity-aware datasets, we propose an \textit{identity-agnostic joint training scheme}. Specifically, in addition to the identity-preservation constraint, we embrace the conventional large-scale face datasets FFHQ~\cite{karras2019style} to guide the encoder to learn a meaningful embedding through reconstruction. In this case, the number of identities perceived by the model increases dramatically, significantly boosting the output quality and the generalizability.
With the aforementioned components, our identity encoder is able to produce \textit{personalized embeddings} with as few as one image, without any test-time optimization.

In addition to generation, our solution enables the possibility of multiple downstream tasks that require personalization. In particular, we demonstrate through experiments our method can be extended to personalized super-resolution and inpainting, achieving enhanced identity preservation without harming output quality. Empirical results show that our method consistently outperforms prior fine-tuning based approach in terms of image quality, identity preservation, and personalized reconstruction, and the outputs of our model is preferred by raters more than $95\%$ of the time compared with the best performing baseline.

As image synthesis becomes ubiquitous, the increasing demand of personalized synthesis is inevitable. There is an urge to develop a practical framework that efficiently generates images with customized identities. We show that with proper designs, it is possible to obtain generalizable embeddings without the needs of time-consuming optimization and storage of multiple models, significantly reducing the burdens in the user end.

\section{Related Work}
\vspace*{-0.05in}
\label{sec:related_work}
\noindent\textbf{Face Generation and Editing.}
The emerge of high-quality datasets~\cite{karras2018progressive,karras2019style,lee2020maskgan} and sophisticated generative models, such as VAEs~\cite{van2017neural,esser2021taming}, GANs~\cite{brock2018large,karras2018progressive,karras2019style,karras2020analyzing,karras2021alias}, and diffusion models~\cite{ho2020denoising,rombach2022high}, has enabled unlimited possibility in the fields of face generation and editing.
In particular, training on the high-quality FFHQ~\cite{karras2019style} dataset, StyleGANs~\cite{karras2018progressive,karras2019style,karras2020analyzing,karras2021alias} obtain unprecedented performance with a novel style-based architecture, where a latent code is used to control the characteristics of the generated faces.
Diffusion models~\cite{ho2020denoising,rombach2022high} synthesize faces through iteratively denoise the intermediate outputs, starting from a standard Gaussian noise. Instead of training with two competing objectives, diffusion models are trained with a single denoising objective, thus generally possessing better training stability.
Editing is achieved usually by manipulating the latent space. For example, StyleGAN-based approaches~\cite{abdal2019image2stylegan,shen2020interfacegan,ling2021editgan,singh2022paint2pix,liu2022delving,tov2021designing,richardson2021encoding,alaluf2021restyle,harkonen2020ganspace} edit the attributes of an image, such as smile and age, by obtaining an editing direction in the latent space. 
Analogous to the StyleGAN paradigm, Diffusion Autoencoder~\cite{preechakul2021diffusion} learns an additional encoder to generate an embedding, which is conditioned by the diffusion model for generation. Similarly, editing can be achieved by manipulating the latent embeddings produced by the encoder. 
In this work, instead of controlling the attributes of faces, we are interested in controlling the \textit{identity} of the generated faces. Specifically, we follow the pipeline of Diffusion Autoencoder and explore the construction of an effective identity-specific latent code for personalized face generation.

\vspace{0.1cm}
\noindent\textbf{Personalized Face Generation.}
Personalized face generation imposes an additional constraint that the model should generate faces of the same identity. This is usually achieved by fine-tuning a pre-trained model with multiple images provided by the user. Specifically, MyStyle~\cite{nitzan2022mystyle} adopts a pre-trained StyleGAN, and then optimizes the object latent code, followed by fine-tuning the model with the latent codes. By using up to hundreds of images for test-time optimization, MyStyle produces promising outputs. 
In this work, we take one step further and consider methods that do not require test-time optimization and need as few as one image for personalized face generation. This is of great value as one can bypass the computation and storage overhead incurred from model fine-tuning, enhancing the practicality of personalalied face generation.

\vspace*{-0.05in}
\section{Approach}
\vspace*{-0.05in}
\label{sec:approach}

In this section, we present the identity encoder for personalized image generation using diffusion model. We first introduce diffusion models and diffusion autoencoder. Next, we provide the problem definition and overview for our approach, followed by the details for learning the identity encoder. Finally, we describe how to adapt the personalized model to conditional generation applications.

\subsection{Preliminary}
\label{sub:preliminary}

Our method is built on top of diffusion model and diffusion autoencoder. This section provides the background of our approach.

\paragraph{Diffusion Models}
Diffusion process is defined as a Markov chain of length $T$ that gradually inserts Gaussian noise to the data with a variance schedule $\{\beta_t\}_{t=1}^T$. Let $\mathbf{x}_t$ be the latent at timestep $t$ of the process, their distributions can be written as
    $$q(\mathbf{x}_t | \mathbf{x}_{t-1}) = \mathcal{N}(\mathbf{x}_t; \sqrt{1 - \beta_t}\mathbf{x}_{t-1}, \beta_t\mathbf{I}).$$
In cases when $\beta_t$ is small, the distribution has the following closed-form solution~\cite{ho2020denoising}:
    $$q(\mathbf{x}_t|\mathbf{x}_0) = \mathcal{N}(\mathbf{x}_t; \sqrt{\bar{\alpha}_t}\mathbf{x}_0, (1 - \bar{\alpha}_t)\mathbf{I}),$$
where $\alpha_t = 1 {-} \beta_t$ and $\bar{\alpha}_t = \prod_{s=1}^t\alpha_s$. Here $\mathbf{x}_0$ is the target data with distribution $q(\mathbf{x}_0)$. To model the target distribution, Ho~\etal~\cite{ho2020denoising} propose to learn a mapping $\epsilon_\theta(\cdot)$ that estimates the noise $\epsilon_t$ added to $\mathbf{x}_0$ to produce $\mathbf{x}_t$. Specifically, the mapping is optimized with the following loss function:
    $$\mathcal{L} = ||\epsilon_\theta(\mathbf{x}_t, t) - \epsilon_t||,$$
where $\epsilon_\theta$ is instantiate by a dense prediction model such as a U-Net.
During inference, samples in the target distribution are obtained through reverse diffusion process that iteratively removes the noise using the trained noise estimator, starting from a pure Gaussian noise $\mathbf{x}_T\sim\mathcal{N}(\mathbf{0}, \mathbf{I})$.

\paragraph{Diffusion Autoencoder}
Mimicking the latent space manipulation in StyleGAN~\cite{karras2018progressive,karras2020analyzing,karras2021alias}, Diffusion Autoencoder~\cite{preechakul2021diffusion} employs an auxiliary encoder to discover a meaningful latent space for image editing in the diffusion model paradigm. Specifically, a learnable encoder is used to produce an embedding as a condition, and the diffusion model acts as a decoder to produce output images given the conditions. The entire autoencoder is jointly trained with the following loss: 
    $$\mathcal{L} = ||\epsilon_\theta(\mathbf{x}_t, t, \mathbf{z}_{\text{sem}}) - \epsilon_t||,$$
where $\mathbf{z}_\text{sem}$ represents the semantic embedding learned by the encoder. For unconditional sampling, an additional diffusion model $\epsilon_\omega(\cdot)$ is trained to sample $\mathbf{z}_{\text{sem}}$.

\begin{figure}[t]
    \centering
    \includegraphics[width=\linewidth]{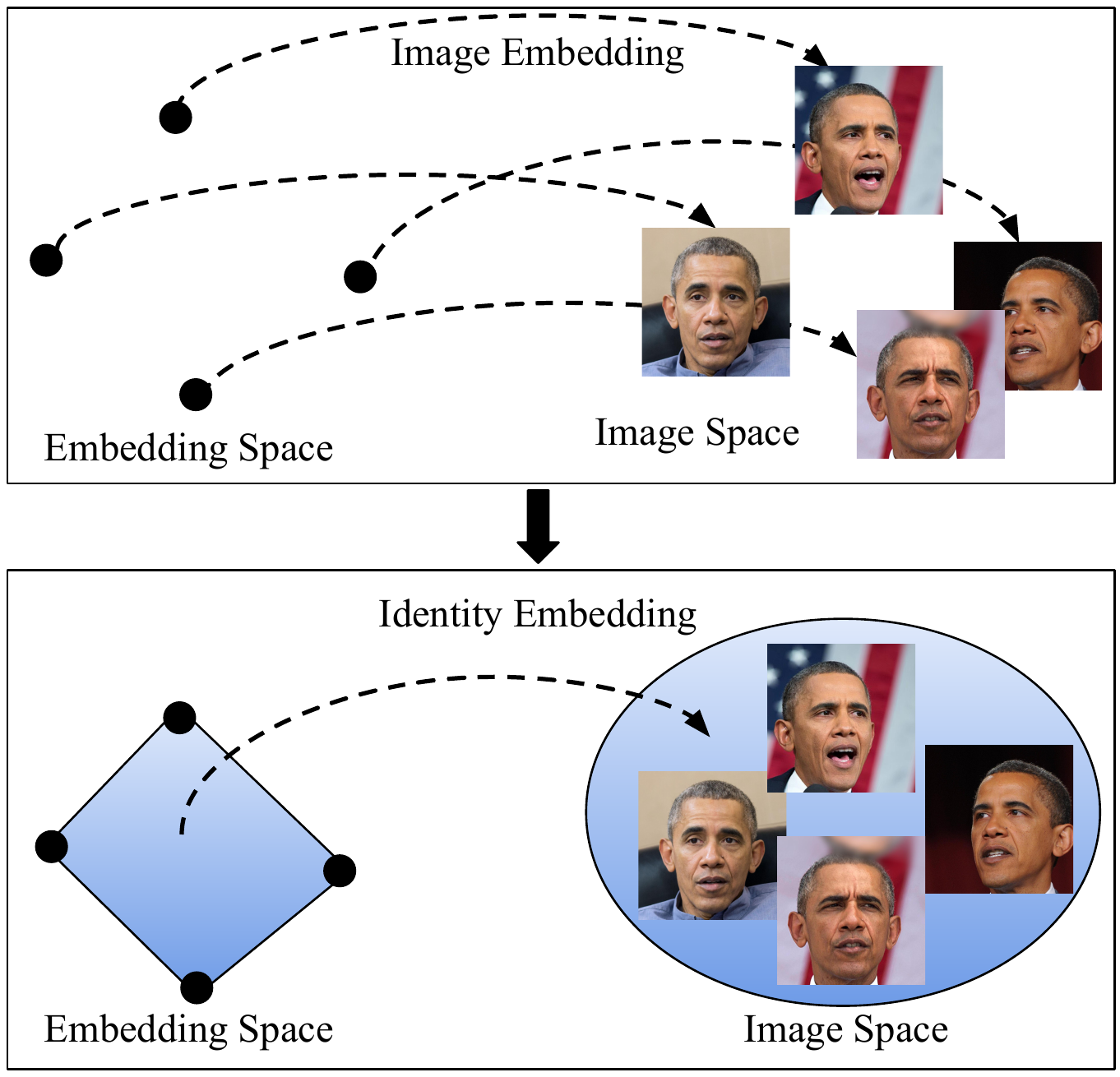}
    \caption{
        Instead of learning an one-to-one mapping between the embedding and image space (top panel), the identity encoder learns a set-to-set mapping. Different combinations of reference images all correspond to the identity representation of the same subject, and each identity embedding maps to all possible images of the subject (bottom panel).
    }
    \vspace{-6pt}
    \label{fig:identity_embedding}
\end{figure}

\subsection{Overview}
\label{sub:overview}

We define our problem as follows. Let $P_{i}$ be the distribution of all images of subject $s_i$ and $Y_{i} = \{\mathbf{y}^{j}_{i} \in P_{i}\}_{j=1}^{N}$ be a set of reference images of the subject. Our goal is to learn an encoder $E$ together with an image generator $G$ such that the generator is able to synthesize images of the same subject conditioned on the identity representation extracted by the encoder from $Y_{i}$:
\begin{equation}
    G(\epsilon, E(Y_i)) \sim P_{i},
\end{equation}
where $\epsilon$ is a random noise. Note that $G$ and $E$ do not depend on the subject, meaning that the same encoder and generator are used to generate images of an arbitrary subject given example images. We further impose two requirements on the model: 1) it should generalize to identities unseen during training, and 2) it should work with a small number of references for each identity (\ie,~$N\,{<}\,10$).

In this work, we adopt a diffusion model as the generator due to its promising results achieved in recent works. In particular, we build the generator and encoder based on Diffusion Autoencoder~\cite{preechakul2021diffusion}, which also learns a diffusion decoder jointly with an image encoder. 

Unlike Diffusion Autoencoder, which conditions only on a single image, our goal is to learn a model conditioned on a set of reference images to improve robustness and capture intra-identity variations. Moreover, the proposed method should allow arbitrary numbers of reference images as condition. 
To this end, we extend Diffusion Autoencoder by aggregating the encoder outputs from multiple images. Specifically, let $Enc$ be an encoder that extracts a feature representation from an image, \ie~$\mathbf{z} = Enc(\mathbf{y})$, we instantiate the encoder $E$ by aggregating the output of $Enc$:
\begin{equation}
    E(Y_i) = \texttt{Aggregate}\left(\{Enc(\mathbf{y}_{i}^{j})\}_{j=1}^{N}\right).
\end{equation}
In this work, we implement the \texttt{Aggregate} operation with average pooling as it allows the model to be generalized to arbitrary numbers of reference images.

\subsection{Identity Encoder}
\label{sub:identity_embedding}
Given the proposed approach, we apply the following training strategies to strike a balance between identity preservation, output diversity, and image quality.

\paragraph{Random average embedding}
In our approach, the mapping between the identity embedding and the output image is essentially a many-to-many mapping, since 1) an identity embedding should correspond to multiple images of a subject with different appearances, and 2) an image could contribute to different embeddings due to the average pooling of different references. See Fig.~\ref{fig:identity_embedding}. Therefore, directly applying Diffusion Autoencoder does not yield good results, as its training process intrinsically leads to a one-to-one mapping. Furthermore, without being trained on average embeddings, the model is unable to produce high-quality outputs when a subject's average embedding changes from one set of reference images to another.

To encourage such a many-to-many mapping, we adopt a random-weighted average embedding as the condition for the generator during training. Specifically, given $N$ images of a person $\{\mathbf{y}_{i}^{j}\}_{j=1}^{N}$, we use
\begin{equation}
    \mathbf{\tilde{z}}_{i} = \sum_{j=1}^N w_{j} \text{Enc}(\mathbf{y}_{i}^{j})
\end{equation}
as the condition embedding for the diffusion model during training. Here $w_{j}\,{\in}\,[0.0, 1.0]$ are randomly sampled weights with the constraint $\sum w_{j}\,{=}\,1$. The model is trained to reconstruct each $\mathbf{y}_{i}^{j}$ using the same embedding $\mathbf{\tilde{z}}_{i}$. In other words, we require that each point within the convex hull of $\{Enc(\mathbf{y}_{i}^{j})\}_{j=1}^{N}$ is a plausible identity embedding for subject $i$, and the diffusion generator should generate all images of the person conditioned on an identity embedding in the convex hull. Hence, we relax the one-to-one constraint in Diffusion Autoencoder.

\begin{figure}[t]
    \centering
    \includegraphics[width=\linewidth]{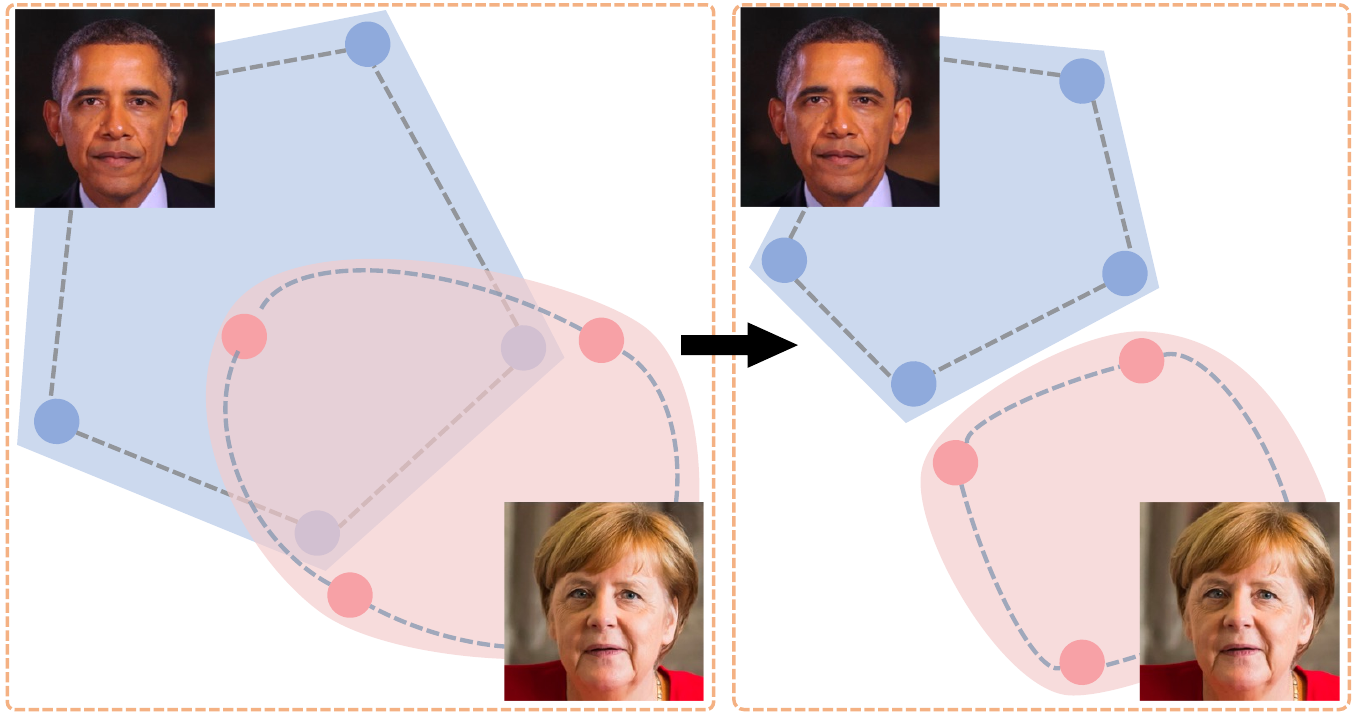}
    \caption{
        To encourage the identity representation to ignore intra-identity variations and focus on inter-identity variations, we impose an identity loss on the identity representation during training. It encourages / discourage the inter-identity / intra-identity representation distances and helps learn a representation that separates different identities. 
    }
    \vspace{-6pt}
    \label{fig:identity_loss}
\end{figure}

\paragraph{Identity Loss}
While training with mean embeddings promotes many-to-many mapping, it could lead to embedding degeneration blurring the boundaries between identities. 
To avoid this caveat, we encourage the encoder to separate different identities in the embedding space as shown in Fig.~\ref{fig:identity_loss}. This is achieved by adding a soft nearest-neighbor loss~\cite{frosst2019analyzing} to the encoder output. 
Specifically, let $\{\mathbf{z}_{k}\}_{k=1}^{K}$ be the encoder output within a batch, and $s_{k}$ be their corresponding identity labels, we minimize the following:
\begin{equation}
    L_{id} = -\frac{1}{K} \sum_{k=1}^{K} \text{log} \frac{\sum_{j \ne k, s_k = s_j} e^{-\frac{\|\mathbf{z}_k - \mathbf{z}_j\|^{2}}{T}}}{\sum_{j \ne k} e^{-\frac{\|\mathbf{z}_k - \mathbf{z}_j\|^{2}}{T}}},
\end{equation}
where $T$ is the temperature controlling the radius of the loss.

\paragraph{Multi-task Learning}
\label{sub:training}
Since both the formation of average embedding and the identity require identity information, the model can only be trained on datasets with identity labels. This greatly limits the amount of available training data, which is amongst the most critical components towards the success of generative models.

To overcome this problem, we apply multi-task learning during training. Specifically, we train the model on images both with and without identity labels.
For images with identity labels, the model is trained with average embeddings and the identity loss described above. 
For images without identity labels, we treat the model as a Diffusion Autoencoder and reconstruct the image conditioned on the embedding of the target image. Our multi-task learning scheme effectively increases the training data size, boosting output quality. Therefore, the final training loss is
\begin{equation}
    L = L_\text{diff}(I_{id}) + \alpha_{1} L_{id} (I_{id}) + \alpha_{2} L_\text{diff} (I_{g}),
\end{equation}
where $L_\text{diff}$ is the standard diffusion loss and $I_{id}$ and $I_{g}$ are images with and without identity information respectively. In practice, half of the images in a training batch are identity-agnostic, and we use $\alpha_{1} = \alpha_{2} = 0.01$.

We hypothesize that reconstruction is an easier task than identity representation extraction, and curriculum learning~\cite{bengio2009curriculum} could be applied to improve the effectiveness of the encoder. To instantiate this, we randomly drop the identity information with probability $p$, with $p$ decreasing linearly from $1.0$ to $0.05$. In this case, the encoder learns meaningful representations through reconstruction, and gradually distill identity information through our delicate training designs.

\subsection{Personalized Conditional Generation}
\label{seb:conditional_generation}
Besides unconditional generation for a subject (\ie,~conditioned only on the identity), many potential applications of generative models require conditional generation. For example, when using generative models for image enhancement like super-resolution, the low-resolution image should be taken as an additional condition. In this work, we introduce an extension of our framework for accommodating extra conditions.

\begin{figure}[t]
    \centering
    \includegraphics[width=\linewidth]{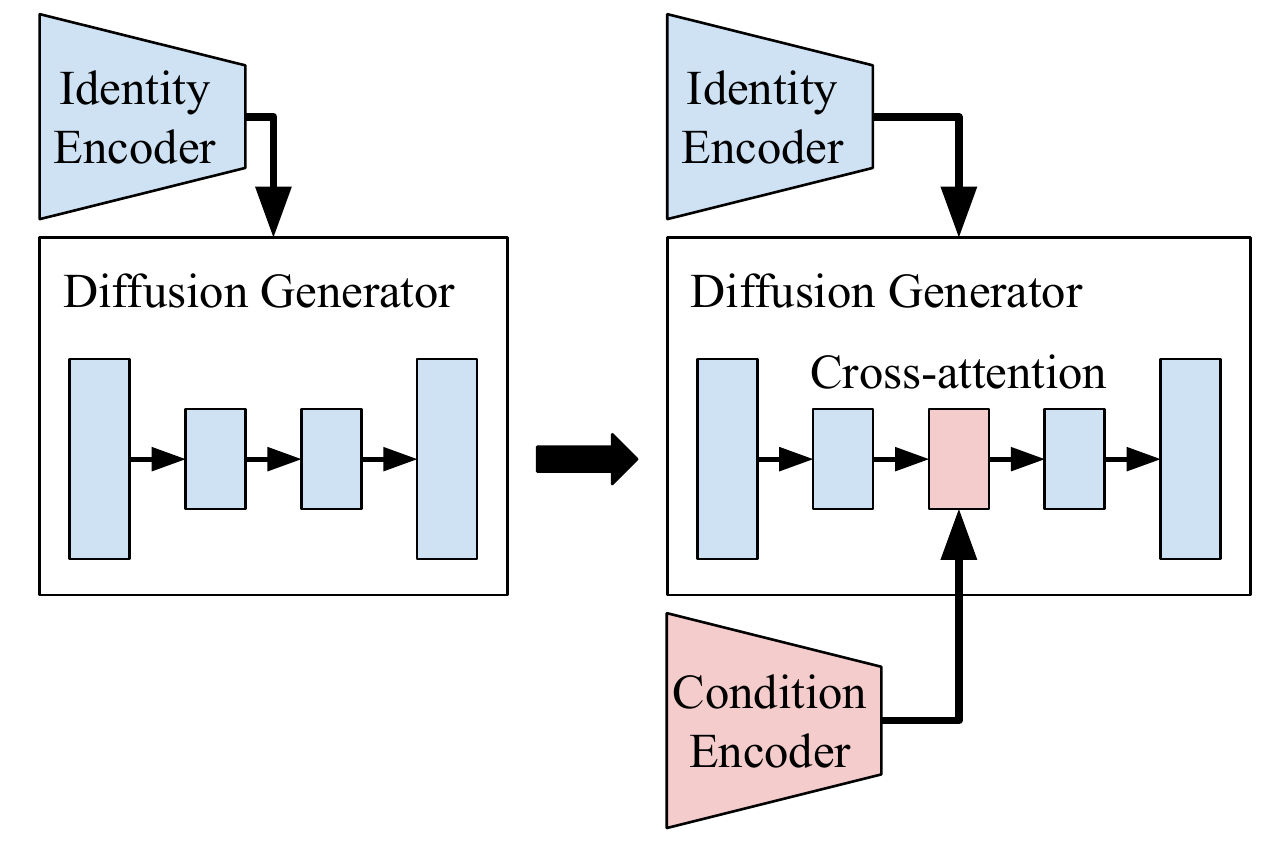}
    \caption{
        The personalized generation model, once learned, can be used to personalize other conditional generation tasks (e.g.~super resolution). This is achieved by injecting the conditional information extracted by a condition encoder into the diffusion generator using cross attention layers. During training for conditional generation, we only update the cross attention layers and the condition encoder.
    }
    \label{fig:conditional_generation}
\end{figure}

We start with a trained personalized generation model. A new condition encoder is used to extract a feature map from the conditional image (or other formats of conditions). The feature map is then injected to the generation model through an attention layer inserted into the diffusion generator. This layer computes the cross-attention between the intermediate feature map of the diffusion generator and the feature map extracted from the conditional image. See Fig.~\ref{fig:conditional_generation} for an illustration. This is similar to the conditioning mechanism of Stable Diffusion~\cite{rombach2022high}. We then train the condition encoder and cross-attention layer while freezing the original personalized generation model.

\vspace*{-0.05in}
\section{Experiments}
\vspace*{-0.05in}
\label{sec:experiments}

\paragraph{Datasets}
To train and evaluate our personalized generation model, we compile a new dataset from existing datasets with identity information:
\begin{itemize}[leftmargin=*,label=$\bullet$,topsep=2pt]
    \setlength{\itemsep}{2pt}
    \setlength{\parskip}{2pt}
    \item \textbf{CelebA}~\cite{liu2015faceattributes} consists 202,599 images from 10,177 celebrities. Among them, 9,564 have more than one image, and the average number of images per identity is 21.
    \item \textbf{CelebRef-HQ}~\cite{li2022learning} consists of 10,555 images from 1,005 celebrities. Each identity has 3--21 images.
    \item \textbf{MyStyle}~\cite{nitzan2022mystyle} consists of 2,154 images from 14 celebrities. Each identity has more than 100 images. The dataset also provides a fixed training and test split, where the average number of test images per identity is 9.
\end{itemize}
We merge the three datasets with the following modifications. First, we remove identities with less than four images. Second, we manually removes identities in \textbf{CelebA} and \textbf{CelebRef-HQ} that overlap with \textbf{MyStyle}. We use \textbf{CelebA}, \textbf{CelebRef-HQ}, and the training split of \textbf{MyStyle} as the training set and the test split of \textbf{MyStyle} as the test set. This leads to the final dataset consisting of 213,487 images from 10,568 identities in the training set and 126 images from 14 identities in the test set\footnote{We do not deduplicate identities between CelebA and CelebRef-HQ.}. Note that our goal is to build a personalized generation model that can generalize to identities not available during training. Therefore, we train our model with and without \textbf{MyStyle} training set to compare the generalizability to new identities.

Besides the datasets with identity information, we also use the \textbf{FFHQ}~\cite{karras2019style} dataset for pre-training. It consists of 70,000 face images without identity information.

\begin{table*}[t]
\small
\centering
\caption{Personalized generation performance. The arrows indicate lowers ($\downarrow$) or higher ($\uparrow$) better for each metric.  Best numbers in \textbf{bold} and second-best in \underline{underlined}.}
\vspace{-6pt}
\begin{tabular}{lccccccccc}
      & \multicolumn{3}{c}{Generation} & \multicolumn{3}{c}{Inpainting} & \multicolumn{3}{c}{Super-resolution}\\
      \cmidrule(lr){2-4} \cmidrule(lr){5-7} \cmidrule(lr){8-10}
      & ID\,$\downarrow$ & FID\,$\downarrow$ & Diversity\,$\uparrow$ & ID\,$\downarrow$ & FID\,$\downarrow$ & LPIPS\,$\downarrow$ & ID\,$\downarrow$ & FID\,$\downarrow$ & LPIPS\,$\downarrow$\\
      \midrule
      DiffAE             & 0.135 & 118.0 & 0.028 & 0.153 & 105.1 & 0.256 & 0.170 & 102.7 & 0.312\\
      MyStyle            & \underline{0.117} & 214.9 & 0.027 & 0.155 & 195.8 & 0.255 & 0.159 & 216.1 & 0.265 \\
      \midrule
      Ours               & 0.119 & \underline{92.9} & \underline{0.115} & \textbf{0.110} & \textbf{94.0} & \textbf{0.125} & \underline{0.118} & \underline{98.6} & \underline{0.146}\\
      Ours (w/ MyStyle)  & \textbf{0.115} & \textbf{91.1} & \textbf{0.125} & \underline{0.111} & \underline{94.9} & \underline{0.129} & \textbf{0.117} & \textbf{97.0} & \textbf{0.143} \\
    \bottomrule
\end{tabular}
\vspace{-3pt}
\label{tab:results}
\end{table*}

\vspace*{-0.05in}
\paragraph{Baselines}
We compare our methods with the following baselines covering both 1) finetuning-based personalized generation model, and 2) diffusion-based autoencoder:
\begin{itemize}[leftmargin=*,label=$\bullet$,topsep=2pt]
    \setlength{\itemsep}{2pt}
    \setlength{\parskip}{2pt}
    \item \textbf{Diffusion Autoencoder (DiffAE)}~\cite{preechakul2021diffusion}---we train a diffusion autoencoder on the personalized dataset. Instead of using the original model during inference, which use the semantic embedding from a single image as the condition for the diffusion decoder, we use the average semantic embedding of the same identity as the condition.
    \item \textbf{MyStyle}~\cite{nitzan2022mystyle}---following the original paper, we finetune a StyleGAN generator~\cite{karras2019style} for each identity using the reference images.
\end{itemize}
For \textbf{DiffAE}, we use the same architecture for the conditional diffusion generator for a fair comparison. Note that unlike \textbf{MyStyle} which assumes that there are hundreds of images for each identity, we target on scenarios where each identity only has a few reference images, i.e.~less than ten, during test time.

\paragraph{Evaluation Metrics}
We evaluate our method on both personalized generation, i.e.~conditioned only on identity, and conditional generation applications. We compare each method using the following metrics for image generation and reconstruction:
\begin{itemize}[leftmargin=*,label=$\bullet$,topsep=2pt]
    \setlength{\itemsep}{2pt}
    \setlength{\parskip}{0pt}
    \item \textbf{Identity score (ID)} computes the distance between the generated images and real images of the given identity. It measures how well the model preserves the identity information and is computed for both personalized generation and conditional generation applications.
    \item \textbf{Fréchet inception distance (FID)}~\cite{NIPS2017_8a1d6947} measures the distribution distance between generated and real images. It measures the quality of generated image images and is computed for both personalized generation and conditional generation applications.
    \item \textbf{Learned Perceptual Image Patch Similarity (LPIPS)}~\cite{zhang2018perceptual} measures the embedding distance between generated and target images using a recognition model. It measures the perceptual similarity between generated and target image and is computed for conditional generation applications.
    \item \textbf{Generation Diversity} is a simple metric we propose to measure the diversity of the outputs. Given multiple outputs, it first computes the mean of each image, and then computes the standard deviation of the numbers. A higher value indicates larger diversity.
\end{itemize}

\vspace{-0.1in}
\paragraph{Implementation Details}
Our model architecture follows that of Diffusion Autoencoder~\cite{preechakul2021diffusion}, except that we replace the attention layers with the implementation of Vision Transformer~\cite{dosovitskiy2020image}. For all diffusion-based methods, we initialize the diffusion generator using a diffusion model pre-trained on FFHQ. The identity encoder is randomly initialized. We train the model using ADAM for 1,000,000 steps with batch size 128 and learning rate $5.0{\times}10^{-5}$. We train the models on $128{\times}128$ image resolution.\footnote{We were unable to train larger models due to resource constraints.} During inference, we run the model for $1,000$ steps using DDPM~\cite{ho2020denoising}. Please refer to the supplementary material for details.

\begin{figure}[t]
    \centering
    \includegraphics[width=\linewidth]{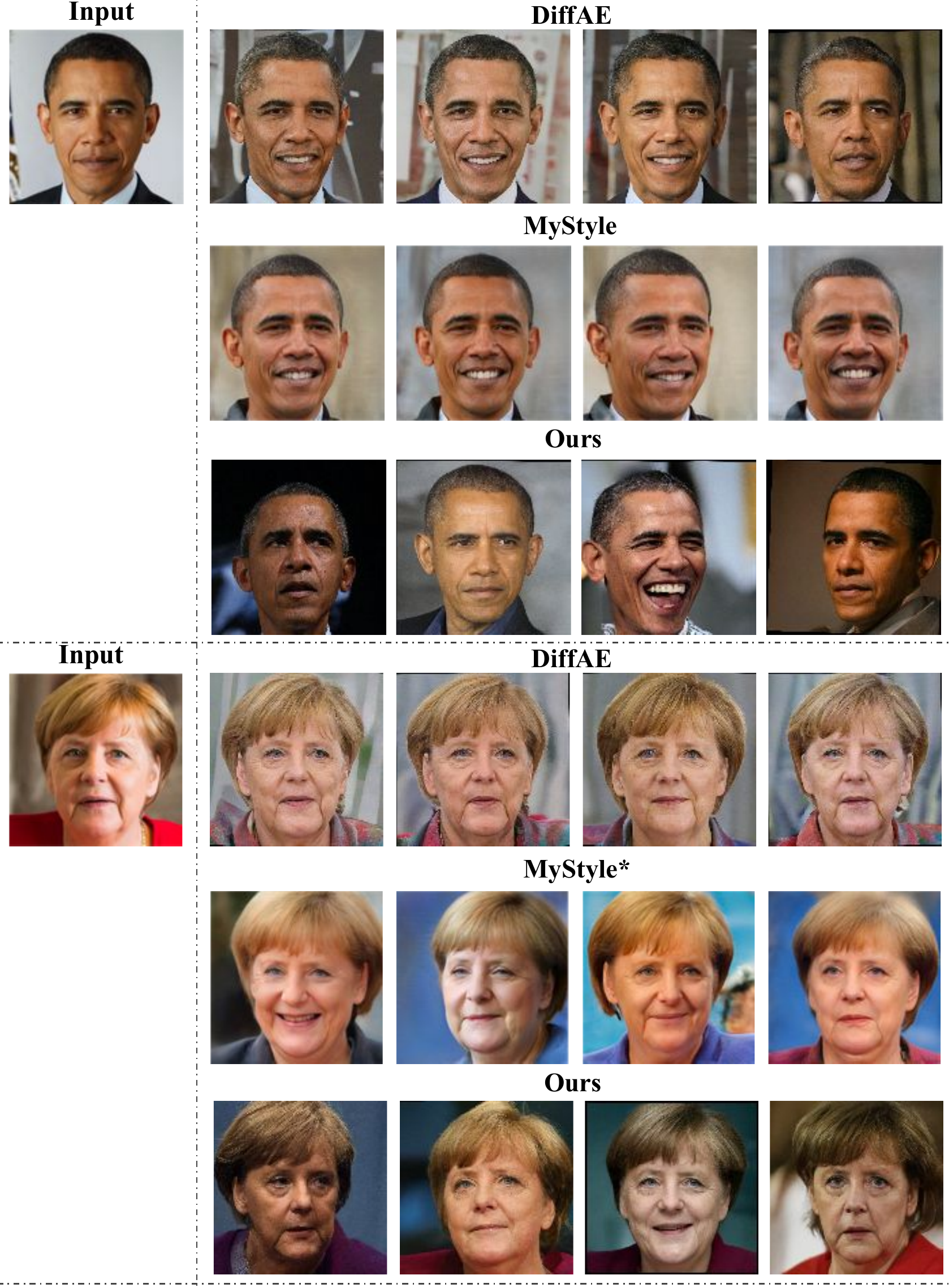}
    \caption{
        Qualitative results on personalized generation. For fair comparison, all methods use eight reference images of the subject. The only exception is MyStyle*, which is trained on the entire training set to demonstrate the necessity of large number of references.
    }
    \label{fig:generation}
    \vspace{-9pt}
\end{figure}

\subsection{Personalized Generation}
\label{sub:exp_generation}

We first evaluate our method on personalized image generation. The results are in Table~\ref{tab:results}. Our method consistently outperforms the baselines in terms of image quality and is on par with the best performing baseline in terms of identity preservation. The results verify the effectiveness of the proposed identity encoder. Note that our method performs similarly when trained with and without MyStyle training set, which demonstrate the generalizability of the identity encoder to unseen identities.

Among the baseline methods, MyStyle achieves the best identity preservation but the worst image quality and output diversity. This shows the limitation of finetuneing-based method, where the performance may drop significantly when there is only limited number of references (less than 15 images in the experiment) for the subject. While DiffAE improves the generated image quality, it also fails to generate diverse outputs, and the identity information is not well preserved. Our identity encoder strikes a better balance between identity preservation and generated image quality.

Fig.~\ref{fig:generation} shows the qualitative examples. The results are consistent with the quantitative metrics, where the outputs of DiffAE and MyStyle lack diversity. As a result, the outputs of the models may not be sufficient to cover all the variations of the subject, which limits their usefulness in potential applications. Note that the output diversity of MyStyle significantly improves when it is trained on the entire MyStyle training set, which shows that it requires a large number of references in order to achieve its best performance. In contrast, our method can generate diverse outputs of the subject from just a few examples. Interestingly, the outputs of DiffAE look like an average image of the subject. This is caused by the fact that DiffAE only learns a point-to-point mapping between the embedding and image space, and the average embedding may not correspond to an real image. The results verify the importance of learning an identity embedding rather than image embedding.

\begin{table}[t]
\small
\centering
\caption{Subjective evaluation results on conditional image generation. We report the percentage of raters that prefer our method.}
\begin{tabular}{lcc}
    & Inpainting (\%) & Super-resolution (\%)\\
    \midrule
    Ours \textit{vs.} DiffAE  & 99.2 & 97.5 \\
    Ours \textit{vs.} MyStyle & 95.8 & 96.7 \\
    \bottomrule
\end{tabular}
\vspace{-9pt}
\label{tab:subjective}
\end{table}

\subsection{Conditional Generation Applications}
\label{sub:exp_applications}

Next, we compare the models on conditional image generation. Specifically, we evaluate the models on 8$\times$ super-resolution and image inpainting. For inpainting, we randomly apply a square mask to the image, where the size of the mask is set to $60\%$ the image size. The results are in Table~\ref{tab:results}. The results are consistent with that of generation, where our methods achieves the best output image quality and reconstruction accuracy. The results further verify the effectiveness of our identity encoder. 

Among the baselines, DiffAE performs particularly worse in terms of reconstruction accuracy, because the reference image dominates the generator due to the design of DiffAE. As a result, DiffAE tends to ignore the additional condition when trained for conditional generation, which leads to a poor reconstruction. MyStyle also performs poorly in terms of reconstruction accuracy, because the outputs of the model are confined by the reference images. When the number of references is small, they may not be sufficient to cover the variations of the subject and limit the output space. As a result, the personalized model is not expressive enough to reconstruct arbitrary images of the subject.

Fig.~\ref{fig:exp_inpainting} shows qualitative results. We can clearly see that DiffAE and MyStyle struggle with authentically reconstructing the inputs, despite that the outputs may look realistic and preserve the identity.  The last row of each panel show the failure cases. For inpainting, our method fail to recover fine-grained attribute, i.e.~the age of the subject. For super-resolution, it fails to generate a realistic face with extreme pose. This may be attributed to the training data, where most of the images contain frontal face, and the model does not generalize well to profile faces.

We further conduct an user study on the conditional generation results. We randomly samples 20 images from the test set for each task and ask the raters to compare the outputs of different methods. For each image, we show the output of the baseline and our method, together with the input and ground truth, and ask raters which method better reconstructs the ground truth. The order of the generated images is randomly permuted to avoid bias. Each image is rated by 6 raters, and the average results are in Table~\ref{tab:subjective}. Our method is preferred by the raters more than $95\%$ of the time, which is consistent with the objective evaluation results.

\begin{figure}[t]
    \centering
    \includegraphics[width=\linewidth]{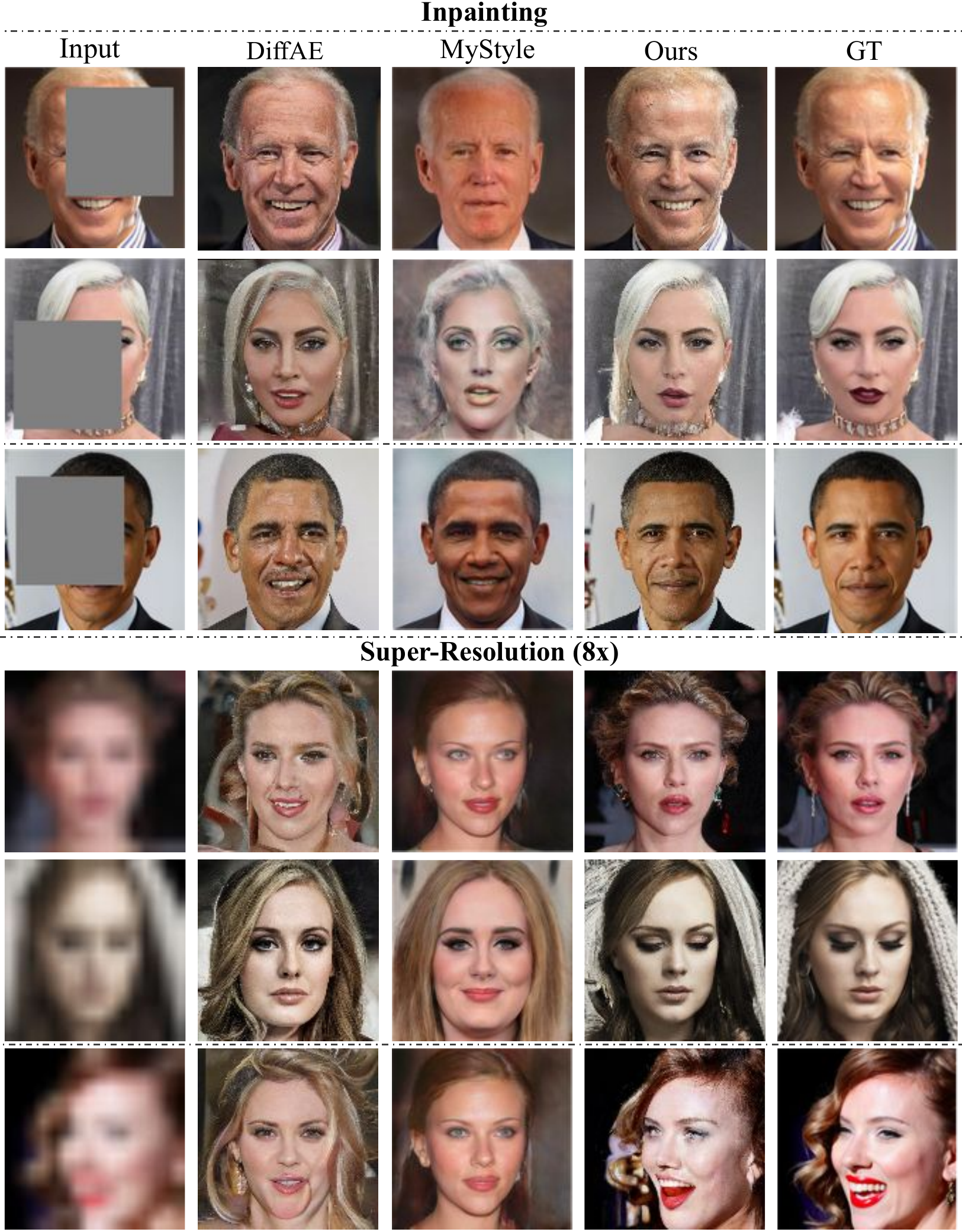}
    \vspace{-12pt}
    \caption{
        Qualitative results for conditional image generation. The last row show failure cases.
    }
    \label{fig:exp_inpainting}
    \vspace{-9pt}
\end{figure}

\vspace*{-0.1in}
\subsection{Ablation Study}
\vspace*{-0.05in}
\label{sub:exp_ablation}

Finally, we conduct ablation study on our method. We first evaluate how each component in our method contributes to the final results. The results are in Table~\ref{tab:ablation}. The results verify the necessity of all three major components in Sec.~\ref{sub:identity_embedding}. While the average embedding leads to worse identity preservation, qualitative results suggest that it is important to generate realistic images from the average embedding of multiple references.

\begin{figure}[t]
    \centering
    \includegraphics[width=\linewidth]{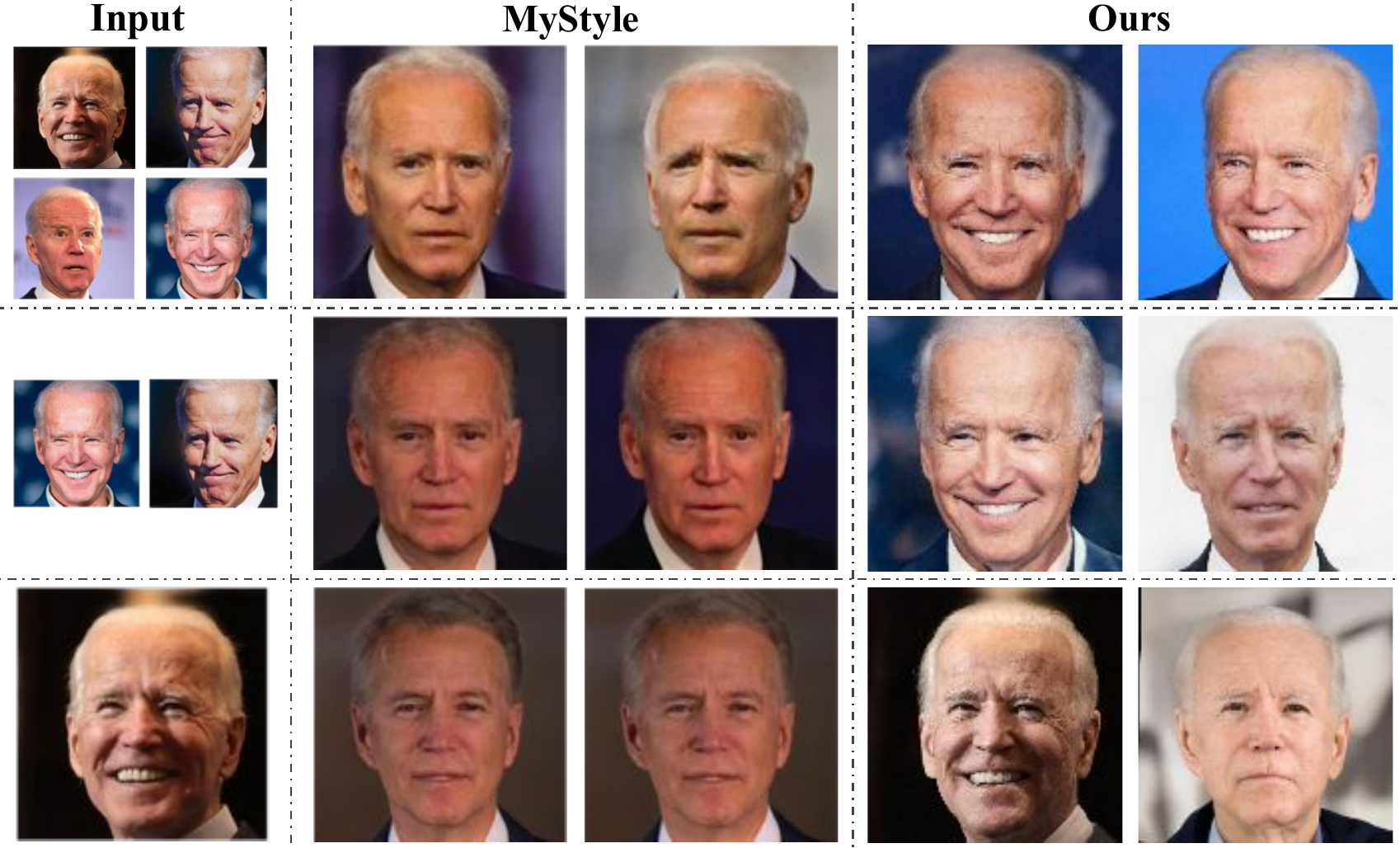}
    \vspace{-12pt}
    \caption{
        Qualitative results showing the effect of number of reference images.
    }
    \vspace{-3pt}
    \label{fig:exp_ablate_face_number}
\end{figure}

Next, we study how the number of references affects the output quality. The results are in Fig.~\ref{fig:exp_ablate_face_number}, were we vary the number of reference images for both our method and MyStyle. We can see that the output quality of MyStyle degrades when the number of references decreases. In contrast, the results of our method remain similar. In fact, our method works with even a single reference. The results show that our method is robust to the number of references and is applicable to more scenarios compared with finetuning-based methods.

\vspace*{-0.05in}
\section{Limitations and Social Impacts}
\vspace*{-0.05in}
\label{sec:discussion}

In this section, we discuss the limitations and potential social impacts of our approach. From the model perspective, one limitation of our approach is that the average embedding strategy may be sub-optimal. Specifically, it may be difficult to capture all potential variations of a subject, and the average embedding may also ``average'' some fine-grained attributes. As the result, the benefit of additional reference may be limited. Future work will explore a more effective method to utilize a large number of references to improve the performance.

Another limitation we observed is that the output quality is subject dependent. A possible reason is the distribution of the training data, which makes the model perform better for highly represented groups. Similar problems has been observed in prior research~\cite{menon2020pulse}. Future work should explore using stronger prior knowledge to build a more robust model. Furthermore, we would like to evaluate the models on larger datasets to study and eliminate potential bias with respect to different demographic groups.

Although personalized image generation is beneficial for many applications, it also increases the risk of abuse. One obvious example is identity forgery, where the model can be used to generate fake images of a person. While prior efforts try to address this problem by detecting images generated by generative model, future work should also consider taking a more active role from the generator side, e.g.~provide a mechanism to recognize images generated by the model.

\begin{table}[t]
\small
\centering
\caption{Ablation study on personalized generation.}
\begin{tabular}{lcc}
    & ID & FID\\
    \midrule
    DiffAE                & 0.135 & 118.0\\
    + average embedding   & 0.146 & 102.2\\
    + multi-task learning & 0.126 & 93.4\\
    + identity loss       & 0.119 & 92.9\\
    \bottomrule
\end{tabular}
\vspace{-9pt}
\label{tab:ablation}
\end{table}

\vspace*{-0.05in}
\section{Conclusion}
\vspace*{-0.05in}

We propose personalized image generation using identity encoder. The identity encoder extracts an identity representation from a set of references of the subject, and a diffusion generator generates new images of the subject conditioning on the identity representation. The same model can generate new images of arbitrary subjects given their reference images. Compared with finetuning-based approaches, our method does not incur additional computation and storage overhead for new subjects, and empirical results show that it achieves better generation quality when there are only a small number of references for each subject.

{\small
\bibliographystyle{ieee_fullname}

}

\section*{Appendices}
\label{sec:appendix}
\appendix
\renewcommand{\thesubsection}{\Alph{subsection}}

The appendix consist of:
\begin{enumerate}[leftmargin=*,label=\Alph*]
    \item Implementation details
    \item Additional qualitative examples
\end{enumerate}

\subsection{Implementation Details}

\begin{figure}[t]
    \centering
    \includegraphics[width=\linewidth]{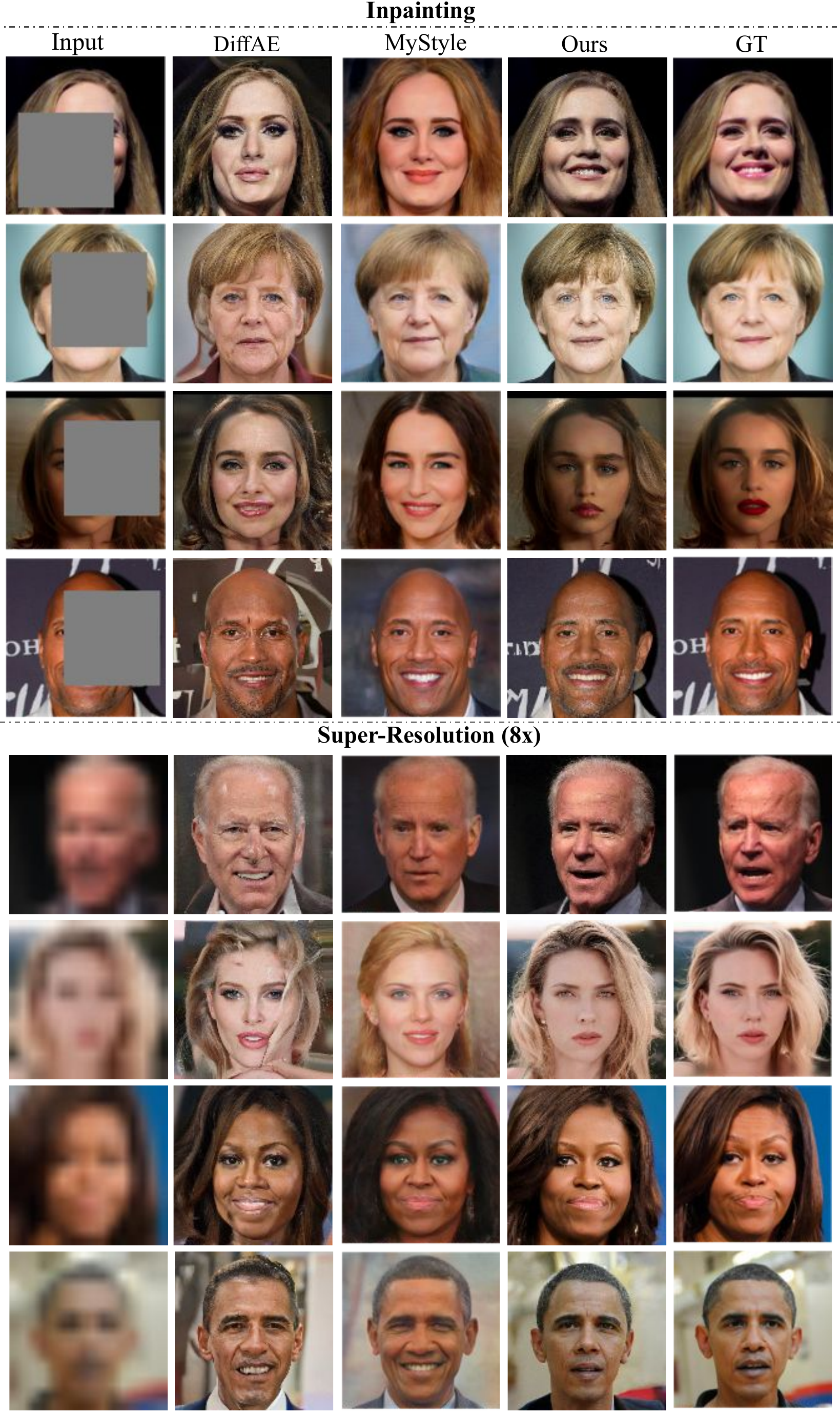}
    \vspace{-12pt}
    \caption{
        Qualitative results for conditional image generation.
    }
    \label{fig:exp_inpainting}
    \vspace{-9pt}
\end{figure}

\begin{figure}[t]
    \centering
    \includegraphics[width=\linewidth]{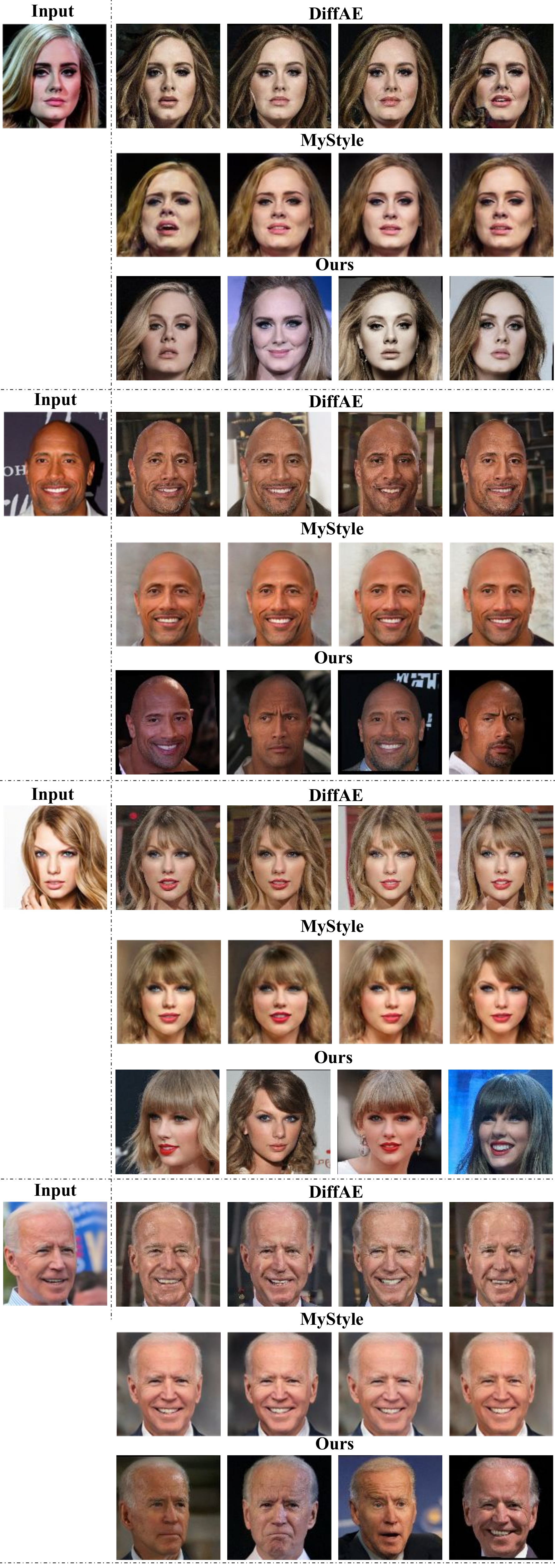}
    \vspace{-12pt}
    \caption{
        Qualitative results for image generation.
    }
    \label{fig:exp_generation}
    \vspace{-9pt}
\end{figure}

In this section, we describe the implementation details that could not fit in the main paper.

\paragraph{Model architecture}
In this paper, we modify the architecture design from DiffAE~\cite{preechakul2021diffusion}, similar to the one used in OpenAI guided-diffusion~\cite{dhariwal2021diffusion}. The diffusion generator is a U-Net of a stacked convolutional encoder and a stacked convolutional decoder. Dense skip connections are in between. For both the encoder and the decoder, the number of base channels is 128 and we set the channel multiplier as $\left[1, 1, 2, 2, 4, 4 \right]$ at resolution $[128^2, 64^2, 32^2, 16^2, 8^2, 4^2]$. In each resolution block, there are two layers each of which contains a timestep-dependent residual module and a self-attention module. Attention is applied at resolution $8\times8\:\text{and}\: 4\times4$ with eight heads. For the identity and conditional encoder, we set the channel multiplier as $\left[1, 1, 2, 2, 4, 4, 4\right]$ at resolution $[128^2, 64^2, 32^2, 16^2, 8^2, 4^2, 2^2]$ and apply attention at resolution $8\times8$, $4\times4$, and $2\times2$. The condition feature map is from the final layer of the encoder.

\paragraph{Model learning and inference}
For training, we set the number of diffusion steps $T$ to 1000. Optimizer AdamW~\cite{loshchilov2017decoupled} is applied with $\beta_1=0.9, \beta_2=0.999$. We keep the learning rate a constant of $lr=5e{-5}$ with $10e^4$ linear warm-up steps. For inference, we run DDPM sampler for 1000 steps. We use cosine $\beta$ scheduler~\cite{nichol2021improved} and set $\beta_{0}=0.0001, \beta_{T}=0.02$.

\subsection{Qualitative Results}

In this section, we present additional qualitative results. Fig.~\ref{fig:exp_generation} and Fig.~\ref{fig:exp_inpainting} extend Fig.~5 and Fig.~6 in the main paper respectively. The results are consistent with those in the main paper, where our method consistently provides better reconstruction accuracy in inpainting and super-resolution and generate more diverse output in unconditional generation.

\end{document}